\documentclass{article}
\usepackage{spconf,amsmath,graphicx}
\usepackage{booktabs} 
\usepackage{pifont}
\usepackage[colorlinks=true, citecolor=black]{hyperref}
\usepackage{amssymb}
\usepackage{amsmath}
\usepackage{threeparttable} 

\title{LVD-GS: Gaussian Splatting SLAM for Dynamic Scenes via Hierarchical Explicit-Implicit Representation Collaboration Rendering}
\name{Wenkai Zhu,  Xu Li$^{*}$, Qimin Xu, Benwu Wang,  Kun Wei,  Yiming Peng and  Zihang Wang \thanks{$^{*}$ is the corresponding author. Email: lixu.mail@163.com. This work was supported in part by the National Key Research and Development Program of China under Grant 2022YFB3904404,in part by the National Natural Science Foundation of China under Grant 62473099.Website: https://zwk0901.github.io/LVD-GS2025.}} 
\address{School of Instrument Science and Engineering, Southeast University, Nanjing, China}
\begin{document}
\ninept
\maketitle
\begin{abstract}
3D Gaussian Splatting SLAM has emerged as a widely used technique for high-fidelity mapping in spatial intelligence. However, existing methods often rely on a single representation scheme, which limits their performance in large-scale dynamic outdoor scenes and leads to cumulative pose errors and scale ambiguity. To address these challenges, we propose \textbf{LVD-GS}, a novel LiDAR-Visual 3D Gaussian Splatting SLAM system. Motivated by the human chain-of-thought process for information seeking, we introduce a hierarchical collaborative representation module that facilitates mutual reinforcement for mapping optimization, effectively mitigating scale drift and enhancing reconstruction robustness. Furthermore, to effectively eliminate the influence of dynamic objects, we propose a joint dynamic modeling module that generates fine-grained dynamic masks by fusing open-world segmentation with implicit residual constraints, guided by uncertainty estimates from DINO-Depth features. Extensive evaluations on KITTI, nuScenes, and self-collected datasets demonstrate that our approach achieves state-of-the-art performance compared to existing methods.
\end{abstract}
\section{Introduction}
\label{sec:intro}

The recent advent of 3D Gaussian Splatting (3DGS) \cite{cheng2025outdoor, GS-SLAM, Gaussian-slam} has enabled high-fidelity photo realistic mapping for autonomous robotic SLAM systems, which is a core technology for embodied intelligence\cite{ren2024embodied, liu2025aligning}. Within this domain, 3D scene representation has emerged as a critical research frontier, driving the development of diverse sparse \cite{KISS-ICP, F-LOAM, FAST-LIVO2} and dense \cite{PIN-SLAM, G²-Mapping, GS-LIVO} representation methodologies that significantly enhance scene understanding.

However, existing 3DGS-SLAM systems exhibit limited performance in complex outdoor scenarios due to reliance on single-representation constraints \cite{GS-SLAM, Gaussian-slam}, facing significant challenges in large-scale dynamic scenes. 
The inherent highly dynamics of outdoor scenes, leading to cumulative errors and trajectory drift \cite{kong2024sc_lpr}, which critically degrades Gaussian point cloud initialization essential for 3DGS performance. Building on prior works \cite{cheng2025outdoor, Gaussian-slam, LiV-GS, Gaussian-Splatting-SLAM} for outdoor 3DGS SLAM, we identify two core challenges: \textbf{the limitations of single-representation constraints} and \textbf{dynamic object interference} . 

On one hand, outdoor scenes provide abundant perceptual cues derived from highly discriminative features in both semantic and appearance domains. However, some existing indoor/outdoor 3DGS SLAM systems rely primarily on pixel-level photometric or geometric reconstruction for optimization \cite{cheng2025outdoor, GS-SLAM, SplaTAM, yang2025opengs,zhu2025loopsplat}. This inherent characteristic leads to \textbf{lack of higher-level semantic representation and global feature understanding} in unbounded outdoor scenes.

On the other hand, due to the highly dynamic nature of outdoor environments, the lack of   \textbf{dynamic modeling} degrades subsequent pose estimation and map reconstruction. Although existing methods attempt to remove these dynamic elements through masking \cite{GaussianEnhancer,xu2024dg,zhou2025dynamic}, they often apply rigid removal strategies without considering the loss of feature consistency during ego-motion and lack fine-grained analysis of dynamic regions.
Therefore, these issues raising the fundamental question: \textbf{how to simulate the human chain-of-thought process to selectively focus on outdoor rich scene information through explicit-implicit representation.}

\begin{figure}
    \centering
    \includegraphics[width=1\linewidth]{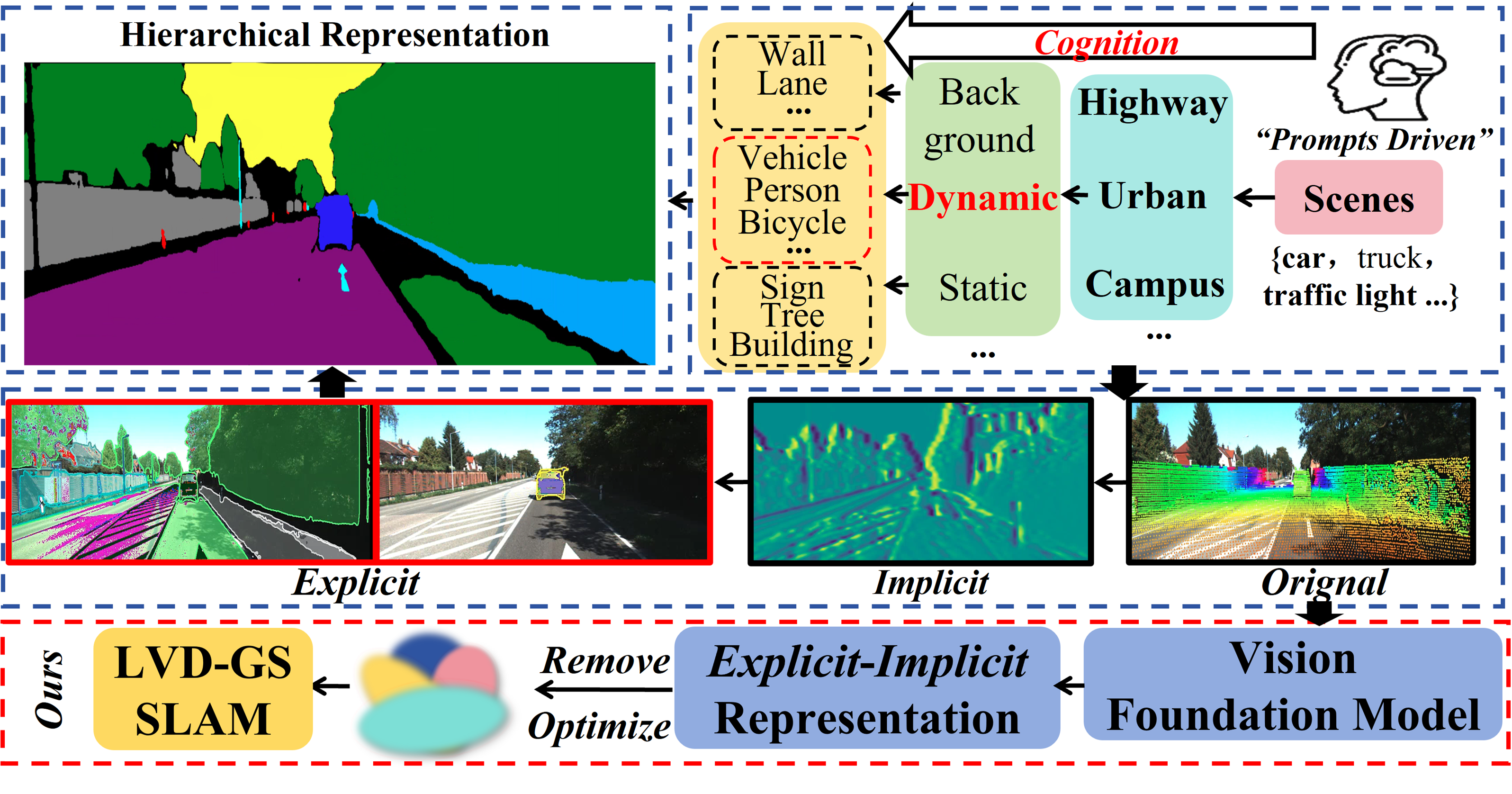}
    \caption{An overview of the chain-of-thought process,  we leverage the high-level semantic understanding  to construct hierarchical explicit-implicit collaborative representation  constraints.}
    \label{intro}
\end{figure}

To address these challenges, we propose  \textbf{LVD-GS} SLAM, a novel LiDAR-Visual Gaussian Splatting SLAM framework designed for dynamic outdoor scenes.  As illustrated in Fig.~\ref{intro}, building on Vision Foundation Models (VFMs), we propose an advanced representation collaboration mechanism that facilitates mutual reinforcement to optimize the mapping process, which effectively resolving scale ambiguity and enhancing reconstruction fidelity. 
Subsequently, we propose a joint dynamic modeling module utilizing open-world segmentation with implicit residual constraints to generate finer-grained dynamic object masks. 
The key innovations and contributions of this paper are highlighted as follows:

\begin{figure*}
    \centering
    \includegraphics[width=1\linewidth]{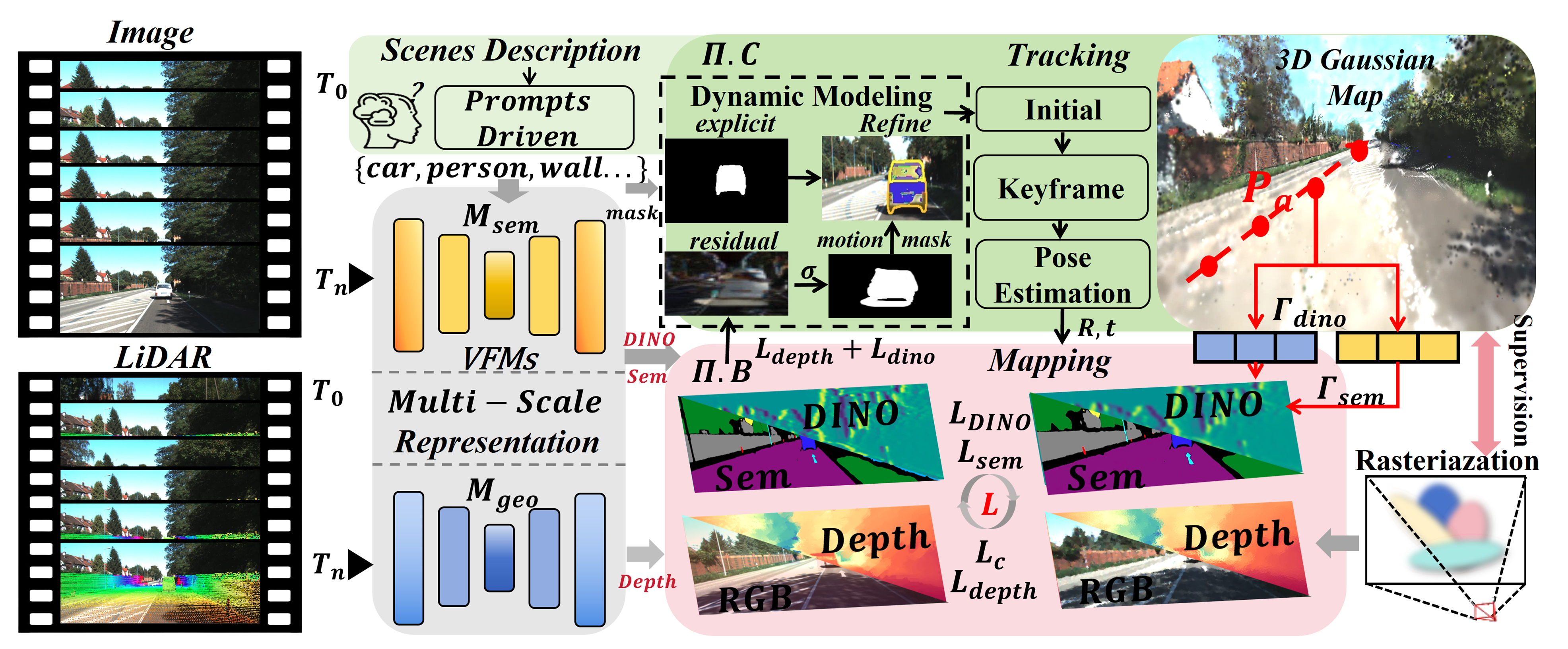}
    \caption{\textbf{SGD-GS SLAM System Overview.} A large-scale 3D Gaussian Splatting framework incorporating a multi-scale representation collaboration module, joint dynamic modeling module. We optimize camera poses using $L$ loss to establish initial pose priors, and refine these poses by incorporating 3D geometric information through scan-to-map registration follows the KISS-ICP\cite{KISS-ICP}. To alleviate memory constraints, the map is partitioned into localized submaps maintained within a fixed spatial range.}
    \label{framework}
\end{figure*}

(1) We propose a novel LiDAR-Visual 3D Gaussian Splatting SLAM framework for dynamic scenes, termed \textbf{LVD-GS}, which incorporates hierarchical representations collaboration of geometric, semantic, and DINO feature to effectively higher-level understanding and achieve high-fidelity reconstruction.

(2) We propose a joint dynamic modeling approach that leverages uncertainty estimation from DINO-Depth features, which combining open-world segmentation with implicit residual constraints to produce fine-grained dynamic object masks.

(3)Extensive evaluations on KITTI, nuScenes, and self-collected datasets demonstrate that our method achieves state-of-the-art performance in both pose estimation accuracy and novel view synthesis among existing 3DGS-SLAM systems.

\section{METHOD}
\label{sec:METHOD}
In this section, we will introduce the  \textbf{LVD-GS} SLAM pipeline, illustrated in Fig.~\ref{framework}. We process RGB frames and LiDAR point clouds using known camera intrinsics $\mathbf{K} \in \mathbb{R}^{3\times3}$. Our framework integrates two core novel modules: (1)  Hierarchical Representation Collaboration Rendering(Sec. 2.1) (2) Explicit-Implicit Joint Dynamic Modeling (Sec. 2.2)

\subsection{Hierarchical Representation Collaboration Mapping}


\subsubsection{Hierarchical Representation Extraction}
we leverage Grounded SAM \cite{Grounding-dino} -equipped with scene-aware prompt generation—to extract semantic   \cite{Semantic-Guided-Gaussian} and DINO features.The depth features are generated through LiDAR point cloud projection onto image planes and  densified using DepthLab \cite{liu2024depthlab}. This integration builds hierarchical Sem-Geo-DINO representations that unify semantic, geometric and appearance attributes across multi-scale spaces, establishing robust consistency constraints.

\subsubsection{Representation Collaboration Rendering}
To enhance the geometric and photometric fidelity of the Gaussian map, we propose a \textbf{Hierarchical Representation Collaboration Rendering Module} optimized using a novel loss function that enforces multi-scale consistency between differentiable renderings and ground truth.

We construct color and depth loss \cite{GaussianEnhancer} by comparing the rendered RGB and depth values with the ground truth values.
\begin{equation}
\begin{aligned}
\mathcal{L}_{c} &= \frac{1}{|\mathcal{M}|} \sum_{i=0}^{|\mathcal{M}|} \left\| C_i - C_i^{gt} \right\|, \\
\mathcal{L}_{depth} &= \frac{1}{|\mathcal{M}|} \sum_{i=0}^{|\mathcal{M}|} \left\| D_i - D_i^{gt} \right\|
\end{aligned}
\end{equation}
where \( {C}_{i},{D}_{i} \) are rendered RGB and depth values, \( {C}_{i}^{gt},{D}_{i}^{gt} \)
are ground truth values.

For supervising semantic information, we employ cross-entropy loss. Notably, during semantic rendering, we detach the gradient to prevent this loss from interfering with the optimization of geometry and appearance features.

\begin{equation}
{\mathcal{L}}_{s} =  - \mathop{\sum }\limits_{{m \in  M}}\mathop{\sum }\limits_{{l = 1}}^{L}{p}_{l}\left( m\right)  \cdot  \log {\widehat{p}}_{l}\left( m\right)
\end{equation}
where \( {p}_{l} \) represents multi-class semantic probability at class
\( l \) of the ground truth map.

To integrate higher-level scene understanding encoded in the features, we introduce a DINO-feature loss: $\mathcal{L}_{dino}$, to guide the optimization of the enriched scene representation. This loss measures the feature similarity between the DINO features $F_i$ and the rendered feature maps $F_i'$:

\begin{equation}
\mathcal{L}_{\text{dino}} = \frac{1}{N_d} \sum_{i=0}^{N_d} \left( 1 - \frac{F_i \cdot F_i'}{\|F_i\|_2 \cdot \|F_i'\|_2} \right)
\end{equation}
where $N_d$ denotes the feature dimension of DINO, and $i$ indexes the feature vectors. 
Finally, the complete multi-scale feature loss function \( \mathcal{L} \) is the weighted sum of the above losses:

\begin{equation}
\mathcal{L} = {\lambda }_{s}{\mathcal{L}}_{s} + {\lambda }_{dino}{\mathcal{L}}_{dino} + {\lambda }_{c}{\mathcal{L}}_{c} + {\lambda }_{depth}{\mathcal{L}}_{depth} 
\end{equation}
where \({\lambda }_{s},{\lambda }_{dino},{\lambda }_{c},{\lambda }_{depth} \) are weighting coefficients.

\subsection{Explicit-Implicit Joint Dynamic Modeling}
\subsubsection{Uncertainty Prediction}

we adapt this approach to outdoor dynamic scenes by modeling per-pixel Gaussian distributions. This uncertainty representation, derived from fused DINO-Depth features,      facilitates joint implicit constraints across geometric and appearance domains. The residuals U are defined as:

\begin{equation}
U = \lambda'_{dino} \mathcal{L}_{dino} + \lambda'_{depth} \mathcal{L}_{depth} 
\end{equation}

We leverage the rapid rendering capability of \textit{3D Gaussian Splatting} (3DGS) to incorporate the residuals \( U \) into an objective function for estimating a per-pixel uncertainty map. This map is subsequently thresholded to generate a binary motion mask \( \mathcal{M}_{implicit}(u) \), which is used to filter dynamic keypoints from keyframes and prevent their incorporation into the map.
\begin{equation}
\mathcal{M}_{implicit} = \mathbb{I}\left(  \underset{\sigma}{\text{min}} \, \frac{1}{HW} \sum_{i=1}^{H} \sum_{j=1}^{W} \rho(U_{ij}, \sigma) \right)
\end{equation}

\subsubsection{Refinement of Dynamic masks}
To enhance the accuracy and completeness of dynamic object segmentation, we introduce an uncertainty-aware joint modeling approach that integrates explicit open-world segmentation with implicit residual constraints. This fusion yields more precise dynamic object masks, formulated as:

\begin{equation}
\mathcal{M}_{refine} = \mathcal{M}_{explicit}  \cap \mathcal{M}_{implicit} 
\end{equation}

\noindent where \( M_{\text{explicit}} \) denotes the mask obtained from open-world segmentation and \( M_{\text{implicit}} \) represents the mask derived from implicit residual constraints.

%

\section{EXPERIMENTS}
\begin{figure}
    \centering
    \includegraphics[width=1\linewidth]{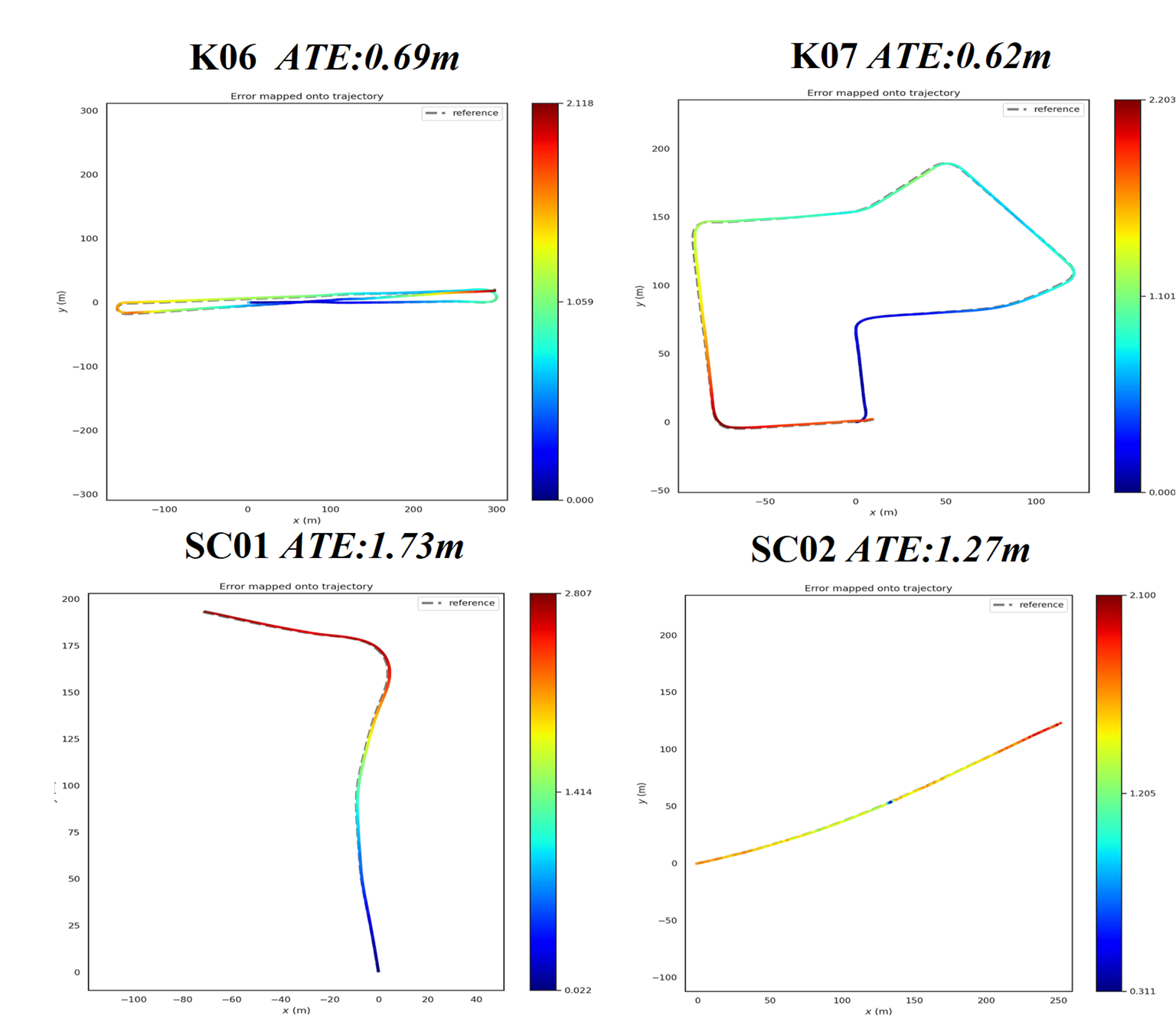}
    \caption{\textbf{Trajectory Visualization.} Due to the memory constraints, other 3DGS-SLAM methods can not run to completion on all sequences, we present only our method's trajectory and error.}
    \label{fig:sy2_pos}
\end{figure}

\begin{table}[t]
\centering
\caption{Pose estimation  performance comparison on KITTI and self-collected datasets. ATE-RMSE is used as the primary metric.
} 
\label{tab:kitti_performance}
\fontsize{9pt}{10.8pt}\selectfont 
\setlength{\tabcolsep}{1.5pt} 
\begin{tabular}{@{}l l c c c c c c c c c@{}}
\toprule
\textbf{Methods} & \textbf{K03} & \textbf{K05} & \textbf{K06} & \textbf{K07} & \textbf{K09} & \textbf{K10} & \textbf{SC01} & \textbf{SC02} \\
\midrule
MonoGS\cite{MONOGS}            & 57.27 & 51.47 & 93.81 & 51.23 & 81.23 & 61.96 & 68.43 & 56.24  \\
SplaTAM\cite{SplaTAM}            & 10.31 & 37.13  & 53.78 & 32.82 & 70.23 & 33.96 & 45.12 & 38.74 \\
OpenGS\cite{yang2025opengs}            & 19.42 & 17.39  & 26.47 & 14.74 & 29.31 & 11.53 & 20.87 & 19.73  \\
S3POGS\cite{cheng2025outdoor}        & 6.36 & 5.94  & 9.34 & 5.63 & 8.64 & 6.52 & 8.63 & 7.12 \\
\midrule
\textbf{Ours}                 & \textbf{1.74}  & \textbf{1.37}  & \textbf{0.69} & \textbf{0.62} & \textbf{2.19} & \textbf{1.45} & \textbf{1.73} & \textbf{1.27}  \\
\bottomrule
\end{tabular}
\end{table}

\begin{figure*}
    \centering
    \includegraphics[width=1\linewidth]{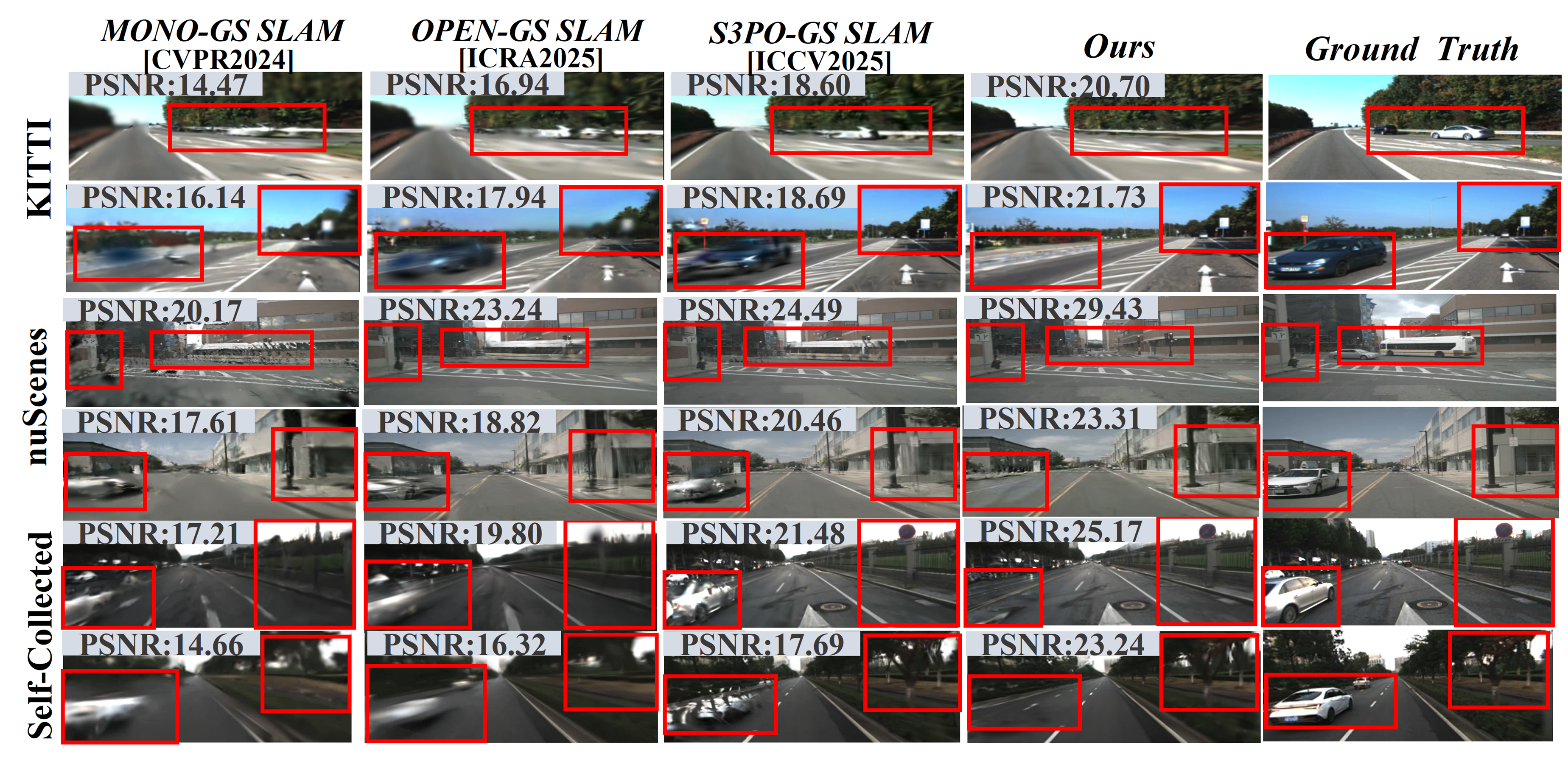}
    \caption{\textbf{Novel view synthesis results on KITTI (top) , nuScenes(mid) and  Self-Collected datasets (bottom)}. Our approach effectively handles complex dynamic environments through a Dynamic Modeling module and Representation Collaboration constraints.}
    \label{fig:nvs}
\end{figure*}

\begin{figure}
    \centering
    \includegraphics[width=1\linewidth]{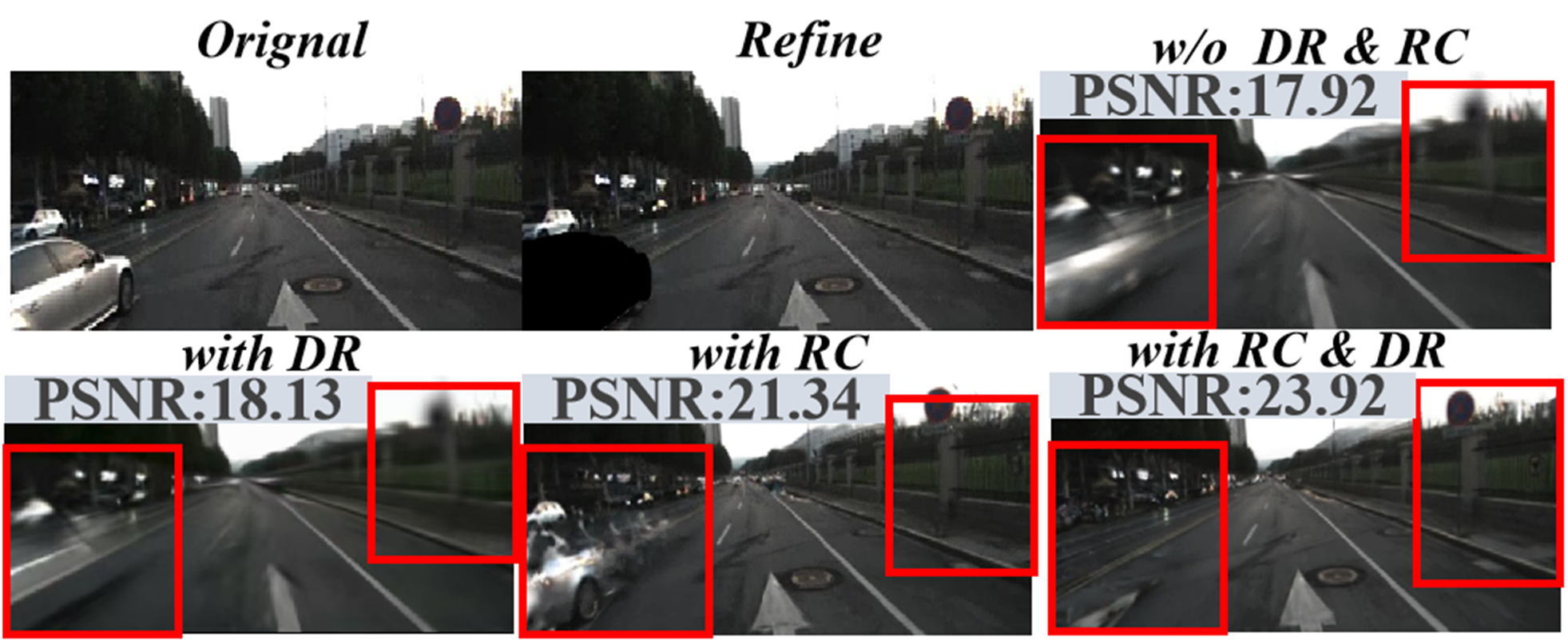}
    \caption{\textbf{Ablation study.} Comparison with two novel modules: Dynamic Modeling and Representation Collaboration.}
    \label{fig:abla_2}
\end{figure}

\subsection{Implementation and Experiment Setup}  

We conduct experiments on the nuScenes\cite{2020nuscenes}, KITTI\cite{kittidataset} and Self-collected Dataset. To evaluate the rendering performance, we use PSNR and SSIM metrics to assess the rendered images. And we use ATE-RMSE(m) to evaluate the pose estimation performance. We compare our method with SLAM approaches five 3DGS SLAM systems MonoGS\cite{MONOGS}, SplaTAM\cite{SplaTAM}, LoopSplat\cite{zhu2025loopsplat}, OPENGS\cite{yang2025opengs}, S3POGS\cite{cheng2025outdoor}. Our implementation is based on the PyTorch framework and tested in NVIDIA RTX3090Ti GPU.

\subsection{Experiment Results}
\subsubsection{Pose Estimation Results}
We evaluate the pose estimation  performance of our method on the KITTI \cite{kittidataset} dataset and  SC dataset containing urban and campus scenes with dynamic objects. As summarized in  Tab. \ref{tab:kitti_performance}, our approach demonstrates superior tracking accuracy across all datasets. By incorporating multi-scale representations and initializing Gaussians from LiDAR points, our system optimizes pose estimation through multi-level features, providing additional constraints that enhance model convergence. Due to memory constraints, other 3DGS-SLAM methods were evaluated only on the first 350 frames per sequence. However, their tracking threads showed large pose estimation errors in outdoor environments, limiting their applicability in real-world large-scale scenes. S3PO-GS\cite{cheng2025outdoor} performs relatively well due to its introduction of pointmap constraints, which effectively mitigate scale drift.

Furthermore, our Hierarchical Representation Collaboration method enhances the camera pose estimation by capturing accurate, rich contextual information, thereby achieving more robust localization. As shown in Fig. \ref{fig:sy2_pos} presents the trajectories of our method on both the KITTI and self-collected datasets, demonstrating its consistent performance across different environments. These results substantiate the overall superiority of the proposed approach.

\subsubsection{Novel View Synthesis}
\begin{table}[t]
\centering
\caption{Novel View Synthesis Results on KITTI, nuScenes and self-collected datasets.Note: \textbf{P} denotes PSNR. \textbf{S} denotes SSIM.}
\label{tab:nvs_results}
\fontsize{9pt}{10.8pt}\selectfont
\setlength{\tabcolsep}{4.5pt} 
\begin{tabular}{@{}l c c c c c c@{}}
\toprule
\textbf{Method} & 
\multicolumn{2}{c}{\textbf{KITTI\cite{kittidataset}}} & 
\multicolumn{2}{c}{\textbf{nuScenes\cite{2020nuscenes}}} & 
\multicolumn{2}{c}{\textbf{SC}} \\
\cmidrule(lr){2-3} \cmidrule(lr){4-5} \cmidrule(lr){6-7}
 & \textbf{P}↑ & \textbf{S}↑ & \textbf{P}↑ & \textbf{S}↑ & \textbf{P}↑ & \textbf{S}↑ \\
\midrule
MonoGS\cite{MONOGS} & 14.30 & 0.441 & 18.58 & 0.709 & 15.76 & 0.627 \\
SplaTAM\cite{SplaTAM} & 14.62 & 0.473 & 18.29 & 0.723 & 16.17 & 0.669 \\
LoopSplat\cite{zhu2025loopsplat} & 16.43 & 0.74 & 23.07 & 0.761 & 18.42 & 0.754 \\
OPENGS\cite{yang2025opengs} & 15.61 & 0.495 & 22.04 & 0.758 & 17.84 & 0.741 \\
S3POGS\cite{cheng2025outdoor} & 19.73 & 0.646 & 24.25 & 0.827 & 21.64 & 0.780 \\
\midrule
\textbf{Ours} & \textbf{21.24} & \textbf{0.81} & \textbf{28.73} & \textbf{0.893} & \textbf{25.43} & \textbf{0.847} \\
\bottomrule
\end{tabular}
\end{table}

As shown in Tab. \ref{tab:nvs_results}, our method achieves state-of-the-art novel view synthesis performance across both datasets. Compared to current 3DGS-based SLAM baselines, PSNR shows significant improvements: +4.48 dB on nuScenes , +1.51 dB on KITTI and +3.79 dB on SC(self-collected). Fig. \ref{fig:nvs}  demonstrates rendered images across three scenarios(urban, highway and compus). For outdoor environments, our approach generates photorealistic reconstructions with enhanced fidelity in vehicle contours, architectural structures, and road surface details. Notably, in highly dynamic regions, our method successfully filters transient objects while maintaining scene consistency, which reduces tracking drift and ensures temporal coherence in synthesized sequences. These results demonstrate the capability of our hierarchical representation collaboration in mitigating scale drift in outdoor scenes and validate the efficacy of the explicit-implicit joint dynamic modeling module in complex urban settings.
\subsection{Ablation Study}
In this section, we evaluate the effectiveness of individual modules within our proposed LVD-GS framework. As summarized in Table~\ref{tab:ablation_results} and illustrated in Fig.~\ref{fig:abla_2}, the Dynamic Modeling and Representation Collaboration components effectively reduce cumulative drift in outdoor environments. We further compare novel view synthesis performance between these two novel modules. Our results show that the Representation Collaboration optimization yields superior performance in large-scale outdoor scenes, where Sem-Geo-DINO cues significantly enhance mapping quality.
\begin{table}[t]
\centering
\caption{Ablation Study on Two Core Modules}
\label{tab:ablation_results}
\fontsize{9pt}{10.8pt}\selectfont
\setlength{\tabcolsep}{4.5pt}
\begin{tabular}{@{}ccc ccc c@{}}
\toprule
\textbf{Dynamic} & \textbf{Representation} & \textbf{PSNR} & \textbf{SSIM} & \textbf{LPIPS} & \textbf{ATE} \\
\textbf{Modeling} & \textbf{Collaboration} & \textbf{(dB)}↑ & ↑ & ↓ & \textbf{(m)}↓ \\
\midrule
\ding{55} & \ding{55} & 20.07 & 0.724 & 0.577 & 10.54 \\
\ding{51} & \ding{55} & 22.79 & 0.780 & 0.513 & 8.42 \\
\ding{55} & \ding{51} & 23.27 & 0.804 & 0.498 & 2.97 \\
\ding{51} & \ding{51} & \textbf{25.43} & \textbf{0.847} & \textbf{0.340} & \textbf{1.27} \\
\bottomrule
\end{tabular}
\vspace{-1mm}
\end{table}

\section{CONCLUSION}
We propose LVD-GS SLAM, a novel LiDAR-visual 3D Gaussian Splatting system that tackles dynamic scenes and scale drift in outdoor environments. Unlike other 3DGS-based SLAM methods, our approach uses representations collaboration to constrain mapping optimization and integrates a joint explicit-implicit module for dynamic object removal. Future work we will futher build instance-level cognitive navigation 3DGS maps.



\renewcommand{\baselinestretch}{0.9}\small\normalsize 
\bibliographystyle{IEEEbib}
\bibliography{strings,refs}

\end{document}